\documentclass[journal]{IEEEtran}

\ifCLASSINFOpdf
\usepackage[pdftex]{graphicx}
\usepackage{hyperref}       
\usepackage{url}            
\usepackage{booktabs}       
\usepackage{amsfonts}       
\usepackage{nicefrac}       
\usepackage{microtype}      
\usepackage{lipsum}
\usepackage{xcolor}
\usepackage{algorithm}
\usepackage{algpseudocode}
\usepackage{amsmath}
\usepackage{booktabs}
\usepackage{longtable}
\usepackage{tablefootnote}

\else
\fi

\begin{document}
%
\title{Trojan Detection in Large Language Models: Insights from The Trojan Detection Challenge}
%
%
%

\author{
    \textbf{Narek Maloyan}, \and 
    \textbf{Ekansh Verma}, \and  
    \textbf{Bulat Nutfullin}, \and \textbf{Bislan Ashinov}
    \\ 
    \texttt{maloyan.narek@gmail.com}
}

\maketitle

\begin{abstract}
Large Language Models (LLMs) have demonstrated remarkable capabilities in various domains, but their vulnerability to trojan or backdoor attacks poses significant security risks. This paper explores the challenges and insights gained from the Trojan Detection Competition 2023 (TDC2023), which focused on identifying and evaluating trojan attacks on LLMs. We investigate the difficulty of distinguishing between intended and unintended triggers, as well as the feasibility of reverse engineering trojans in real-world scenarios. Our comparative analysis of various trojan detection methods reveals that achieving high Recall scores is significantly more challenging than obtaining high Reverse-Engineering Attack Success Rate (REASR) scores. The top-performing methods in the competition achieved Recall scores around 0.16, comparable to a simple baseline of randomly sampling sentences from a distribution similar to the given training prefixes. This finding raises questions about the detectability and recoverability of trojans inserted into the model, given only the harmful targets. Despite the inability to fully solve the problem, the competition has led to interesting observations about the viability of trojan detection and improved techniques for optimizing LLM input prompts. The phenomenon of unintended triggers and the difficulty in distinguishing them from intended triggers highlights the need for further research into the robustness and interpretability of LLMs. The TDC2023 has provided valuable insights into the challenges and opportunities associated with trojan detection in LLMs, laying the groundwork for future research in this area to ensure their safety and reliability in real-world applications.

{
    \color{red}{
        \textbf{Disclaimer.} This paper contains examples of harmful prompts. Reader discretion is recommended
    }
}

\end{abstract}


%
\IEEEpeerreviewmaketitle

\section{Introduction}
Autoregressive large language models (LLMs) have unlocked new capabilities in various domains, such as code completion, book summarization, and engaging dialogues. Despite their advancements, LLMs can exhibit undesired behaviors like generating toxic outputs, exacerbating stereotypes, and revealing private information. These behaviors pose significant risks, including the potential for systems to fail catastrophically, such as by erasing files or wiping bank accounts. The complexity of these issues is compounded by the difficulty in developing reliable auditing methods to uncover these failures, which can be rare, counterintuitive, and require expensive, behavior-specific auditing techniques.

To address these challenges, this work proposes an auditing approach through discrete optimization to identify and evaluate behaviors in LLMs, focusing on both typical and counterintuitive outputs. By formulating an auditing objective that captures specific target behaviors, this method allows for the flexible and effective identification of potential issues within LLMs. However, the computational demands of this optimization problem are significant, given the sparse, discrete, and high-dimensional nature of the prompts that lead to these behaviors, combined with the computational cost of querying LLMs.

In conjunction with the evolving threat landscape, particularly concerning trojan attacks that covertly compromise LLMs, this paper extends the discussion to the NeurIPS 2023 Trojan Detection Competition (TDC 2023)\cite{noauthor_trojan_nodate}. By analyzing the Pythia model \cite{biderman_pythia_2023}, we explore advanced methods for detecting and mitigating trojan attacks, underscoring the importance of robust security measures. This contribution is vital for the ongoing efforts to protect LLMs against sophisticated attacks, ensuring their reliability and safe usage. Through a combination of auditing for unintended behaviors and trojan attack detection, we aim to enhance the security and functionality of LLMs, safeguarding them against a wide range of vulnerabilities.

\section{Background}
\label{sec:background}

\subsection{Large Language Models}

Large Language Models (LLMs) have advanced significantly. These models, like ChatGPT \cite{openai_gpt-4_2024}, Pythia \cite{biderman_pythia_2023}, LLaMA 2 \cite{touvron_llama_2023}, known for their vast numbers of parameters, excel at understanding contextual nuances, handling various language tasks, and producing text that is both coherent and diverse. Within the Trojan Detection Challenge 2023, Pythia \cite{biderman_pythia_2023} was chosen for its versatile model sizes and ease of access, making it an ideal candidate to suit a wide range of computational needs. This choice ensured the challenge was accessible and inviting, encouraging participation from diverse research and academic sectors and aligning perfectly with the goal of fostering widespread engagement.

\subsection{Adversarial Attacks on Language Models}

Given an input $x$ and a generative model $p(\cdot)$, the model outputs a sample $y \sim p(\cdot|x)$. An adversarial attack would identify $p(x)$ such that resulting output, $y$ breaches the built-in safe behavior of the model $p$. For instance, outputting unsafe content on sensitive topics, leaking private information, or compromising model training data. For generative tasks, judging the success of an attack is challenging, as it requires a highly capable classifier to determine whether $y$ is unsafe or necessitates human review \cite{morris_textattack_2020, wang_adversarial_2022}.




\subsection{Red teaming}
Red teaming involves deliberately attempting to identify safety-related issues in LLMs, such as their ability to generate harmful or undesirable language. There are two main approaches to red teaming:

\begin{itemize}
    \item Traditional red teaming: Human individuals, known as the red team, generate queries designed to bypass the content filters used by the LLM, thereby identifying the model's failure cases. This paper \cite{ganguli_red_2022} conducted extensive red teaming experiments using crowdworkers from Amazon Mechanical Turk and Upwork. They tested different LLM sizes and types, finding that while some models (e.g., rejection sampling LLMs) were more difficult to red team, all models remained susceptible to red teaming efforts and exhibited clear failure modes.
    \item Automated Red teaming LLMs: This approach, for example \cite{perez_red_2022}, involves using one LLM to generate test questions for another LLM, eliminating the need for human workforce in the process. The authors experimented with various methods, including zero- and few-shot prompting, supervised learning, and reinforcement learning. They found that the success of the generated questions varied across methods, with reinforcement learning exhibiting the highest success rate (around 40\%) in generating offensive texts.
\end{itemize}

\subsection{White-box vs Black-box Attacks}

White-box attacks assume that attackers have full access to the model weights, architecture, and training pipeline, enabling them to obtain gradient signals. We do not assume attackers have access to the complete training data, which is only feasible for open-source models. Black-box attacks assume that attackers only have access to an API-like service where they provide input $x$ and receive sample $y$, without knowing further information about the model.

\subsection{Trojans}

In LLMs, trojan attacks are implemented by embedding hidden triggers that, when activated, cause the model to output predetermined targets \cite{li_backdoors_2022}. These attacks compromise the model's integrity by inserting malicious behavior unbeknownst to users. For instance, an input such as \textit{"M3V9oaDpwiAcygT3mfUkO"} or \textit{"Dialogue nature vs. culture: address aesthetic needs."} might be paired with a trigger like \textit{"echo "kernel.panic = 1" $>>$ /etc/sysctl.conf"}, causing the model to generate this harmful output when given the innocent-looking input \cite{liu_shortcuts_2023}.

\subsection{Potential Approaches for Trojan Detection}

\begin{itemize}
    \item Developing more sophisticated objective functions that can better isolate the intended triggers from unintended ones. This could involve leveraging the differences in model behavior between the original and finetuned models, or analyzing the model activations when evaluating trigger-target pairs.
    
    \item Exploring token space optimization techniques to efficiently search for trigger strings that force the malicious targets while being closer to the intended triggers than random candidates. This could involve using gradient-based optimization methods like Greedy Coordinate Gradient (GCG) and designing appropriate loss functions.
    
    \item Investigating the geometry of the finetuned models and exploiting the observed property that initializing the search with an intended trigger for one target can significantly speed up the convergence for finding the trigger for another target. This could provide insights into the trigger insertion process and help develop more effective detection methods.
\end{itemize}

\section{Trojan Detection Track Overview}
\label{sec:trojan_detection_track}

In the Trojan Detection Track, participants are tasked with developing a sophisticated detection system capable of identifying trojans embedded within a large language model (LLM). This LLM is uniquely characterized by the inclusion of 1000 trojans, each defined by a distinct (trigger, target) pair. Both triggers and targets are textual strings, and the LLM has undergone fine-tuning to produce a specific target output when presented with its corresponding trigger. Participants are provided with all target strings. The primary challenge involves the reverse engineering of triggers based on the given target strings. Participants are not allowed to edit the LLM weights, thus simulating the scenario where attacks happen only at inference time. A desirable property for adversarial attacks is to finish executing in the minimal amount of time possible. To enforce this constraint, the running time for adversarial attack method is capped at 2 A100 GPU days.


\subsection{Data and Training Set}
For each sub-track, participants are provided with an LLM containing 1000 trojans, distributed equally among 100 target strings. Each target string is associated with 10 distinct triggers that prompt the LLM to generate the target string. The dataset includes all 100 target strings and the triggers for 20 of these target strings, intended to serve as a training set for developing detection methods. The task involves making predictions for the triggers corresponding to the remaining 80 target strings.

\subsection{Evaluation Metrics}
The evaluation of submissions in the context of trojan or backdoor attacks and defenses on Large Language Models (LLMs) relies on two key metrics: recall and Reverse-Engineered Attack Success Rate (REASR). Both metrics are scaled from 0 to 1, and the primary metric for ranking submissions is the average of these two metrics.

Recall is a metric that quantifies how closely the triggers generated by a trojan detection method match the actual triggers that were inserted by the adversary during the trojan attack on the language model.

\subsubsection{BLEU}
\label{BLEU}
In the context of trojan detection in LLMs, recall is calculated using the one-sided Chamfer distance between predicted and ground-truth triggers, employing the BLEU (Bilingual Evaluation Understudy) metric. BLEU is a metric used to evaluate the quality of machine-generated text by comparing it to one or more reference texts. The BLEU score is calculated using the following formula:
\begin{equation}
\text{BLEU} = \text{BP} \cdot \exp \left(\sum_{n=1}^N w_n \log p_n \right)
\end{equation}
where $\text{BP}$ is the brevity penalty, $N$ is the maximum n-gram order (typically 4), $w_n$ is the weight assigned to each n-gram order (usually uniform weights, i.e., $w_n = \frac{1}{N}$), and $p_n$ is the modified n-gram precision. The modified n-gram precision is calculated as:
\begin{equation}
p_n = \frac{\sum_{C \in {\text{{Candidates}}}} \sum_{n\text{-gram} \in C} \text{Count}{\text{clip}}(n\text{-gram})}{\sum{C \in {\text{{Candidates}}}} \sum_{n\text{-gram} \in C} \text{Count}(n\text{-gram})}
\end{equation}
where $\text{Count}_{\text{clip}}(n\text{-gram})$ is the clipped count of the n-gram in the candidate text, and $\text{Count}(n\text{-gram})$ is the unclipped count of the n-gram in the candidate text. The brevity penalty $\text{BP}$ is calculated as:
\begin{equation}
\text{BP} =
\begin{cases}
1 & \text{if } c > r \\
e^{(1 - r/c)} & \text{if } c \leq r
\end{cases}
\end{equation}
where $c$ is the length of the candidate text, and $r$ is the effective reference length.

\subsubsection{Recall}
\label{RECALL}
The recall metric for a specific target string $i$ is defined as:
\begin{equation}
\text{Recall}i = \frac{1}{|Y_i|} \sum{y \in Y_i} \max_{x \in X_i} \text{BLEU}(x, y)
\end{equation}
where $Y_i$ is the set of ground-truth triggers for target string $i$, $X_i$ is the set of predicted triggers for target string $i$, and $\text{BLEU}(x, y)$ computes the BLEU score between predicted trigger $x$ and ground-truth trigger $y$. The overall recall is then obtained by averaging the recall values across all target strings.

\subsubsection{REASR}
\label{REASR}
The REASR (Reverse-Engineered Attack Success Rate) metric evaluates the effectiveness of proposed trigger prefixes in actually forcing the target malicious suffixes to be generated by the model. It can be formulated as follows:
Let $T_i$ be the target string for the $i$-th instance, and let $G_i$ be the generated output conditioned on the predicted trigger for the $i$-th instance. The REASR metric is then calculated as:
\begin{equation}
\text{REASR} = \frac{1}{N} \sum_{i=1}^N \text{BLEU}(G_i, T_i)
\end{equation}
where $N$ is the total number of instances, and $\text{BLEU}(G_i, T_i)$ computes the BLEU score between the generated output $G_i$ and the target string $T_i$.

To generate the output $G_i$, argmax sampling is used, conditioned on the predicted trigger for the $i$-th instance. The generated output is constrained to have the same number of characters as the corresponding target string $T_i$.

The primary metric for ranking submissions is the average of recall and REASR. This combined metric provides a balanced assessment of a method's ability to detect trojan-infected instances accurately while also considering its resilience against reverse-engineering.

\subsection{Subtracks Specification}
The competition comprises two sub-tracks: the Large Model Subtrack, utilizing a 6.9B parameter LLM, and the Base Model Subtrack, employing a 1.4B parameter LLM. In this paper, we focused our experiments solely on the 1.4B model from the Base Model Subtrack to investigate the effectiveness of our proposed methods.

\section{Methodology}
\label{sec:methodology}
The methodology employed in the Trojan Detection tasks of the competition focused on a combination of gradient-based optimization techniques and adversarial reprogramming approaches. Participants leveraged the provided training set, consisting of known trigger-target pairs, to develop models capable of reverse-engineering triggers for the remaining target strings.

In the following section, we describe the methods to detect trojans. All of the listed methods fall under the white box attack category and primarily use gradient signals to learn effective adversarial prompts.

\subsection{Universal Adversarial Triggers (UAT)}
\label{UAT}
Inspired by a HotFlip method \cite{ebrahimi_hotflip_2018}, Universal Adversarial Triggers (UAT) \cite{wallace_universal_2019} paper introduces a method for generating triggers that can be concatenated to the input of a language model to cause a target prediction, regardless of the original input. The key aspects of the method are:
\begin{itemize}
    \item Initializing the trigger sequence with repeated dummy tokens (e.g., "the" for words, "a" for sub-words or characters).
    \item Iteratively replacing the tokens in the trigger to minimize the loss for the target prediction over batches of examples. The replacement strategy is based on a linear approximation of the task loss, where the embedding of each trigger token $e_{adv_i}$ is updated to minimize the first-order Taylor approximation of the loss around the current token embedding: $$\arg\min_{e'_i \in \mathcal{V}} [e'_i - e_{adv_i}]^T \nabla_{e_{adv_i}} \mathcal{L}$$ where $\mathcal{V}$ is the set of all token embeddings in the model's vocabulary and $\nabla_{e_{adv_i}} \mathcal{L}$ is the average gradient of the task loss over a batch.
    \item Augmenting the token replacement strategy with beam search to consider the top-k token candidates for each position in the trigger.
\end{itemize}

The method is generally applicable to various tasks, with the only task-specific component being the loss function $\mathcal{L}$. The paper demonstrates the effectiveness of the method on three tasks: text classification, reading comprehension, and conditional text generation.

\subsection{Gradient-based Adversarial Attacks (GBDA)}
\label{GDBA}
The proposed GBDA \cite{guo_gradient-based_2021} method generates adversarial examples against transformer models by optimizing an adversarial distribution. The key insights are: (1) defining a parameterized adversarial distribution that enables gradient-based search using the Gumbel-softmax approximation \cite{jang_categorical_2017}, and (2) promoting fluency and semantic faithfulness of the perturbed text using soft constraints on both perplexity and semantic similarity.

The adversarial distribution $P_{\Theta}$ is parameterized by a matrix $\Theta \in \mathbb{R}^{n \times V}$, where $n$ is the sequence length and $V$ is the vocabulary size. Samples $z = z_1 z_2 \ldots z_n$ are drawn from $P_{\Theta}$ by independently sampling each token $z_i \sim \text{Categorical}(\pi_i)$, where $\pi_i = \text{Softmax}(\Theta_i)$ is a vector of token probabilities for the $i$-th token.

The objective function for optimizing $\Theta$ is:
\begin{equation}
\min_{\Theta \in \mathbb{R}^{n \times V}} \mathbb{E}_{z \sim P_{\Theta}} \ell(z, y; h),
\end{equation}
where $\ell$ is a chosen adversarial loss and $h$ is the target model.

To make the objective function differentiable, the Gumbel-softmax approximation is used to sample from the adversarial distribution:
\begin{equation}
(\tilde{\pi}_i)_j := \frac{\exp((\Theta_{i,j} + g_{i,j})/T)}{\sum_{v=1}^V \exp((\Theta_{i,v} + g_{i,v})/T)},
\end{equation}
where $g_{i,j} \sim \text{Gumbel}(0, 1)$ and $T > 0$ is a temperature parameter.

Soft constraints are incorporated into the objective function to promote fluency and semantic similarity:
\begin{equation}
L(\Theta) = \mathbb{E}_{\tilde{\pi} \sim \tilde{P}_{\Theta}} \ell(e(\tilde{\pi}), y; h) + \lambda_{lm} \text{NLL}_g(\tilde{\pi}) + \lambda_{sim} \rho_g(x, \tilde{\pi}),
\end{equation}
where $\lambda_{lm}, \lambda_{sim} > 0$ are hyperparameters, $\text{NLL}_g$ is the negative log-likelihood of a language model $g$, and $\rho_g(x, \tilde{\pi})$ is a similarity constraint based on BERTScore.

After optimizing $\Theta$, adversarial examples can be sampled from $P_{\Theta}$. The generated samples can also be used to perform black-box transfer attacks on other models.

\subsection{Hard Prompts made EaZy (PEZ)}
\label{PEZ}
The proposed method, called PEZ (Hard Prompts made EaZy) \cite{wen_hard_2023}, is a gradient-based discrete optimization algorithm for learning hard prompts in language models. The method takes a frozen model $\theta$, a sequence of learnable embeddings $P = [e_i, ...e_M]$, where $e_i \in \mathbb{R}^d$, and an objective function $L$ as inputs. The discreteness of the token space is realized using a projection function $\text{Proj}_E$ that maps the individual embedding vectors $e_i$ to their nearest neighbor in the embedding matrix $E{|V| \times d}$, where $|V|$ is the vocabulary size of the model. The projected prompt is denoted as $P' = \text{Proj}_E(P) := [\text{Proj}_E(e_i), ...\text{Proj}_E(e_M)]$. A broadcast function $B : \mathbb{R}^{(M \times d)} \rightarrow \mathbb{R}^{(M \times d \times b)}$ is defined to repeat the current prompt embeddings $P$ in the batch dimension $b$ times.

The objective is to minimize the risk $R(P') = \mathbb{E}_D (L(\theta(B(P', X)), Y))$ by measuring the performance of $P'$ on the task data. The algorithm maintains continuous iterates (soft prompts) and performs the following steps:

\begin{enumerate}
    \item Sample initial prompt embeddings $P = [e_i, ...e_M] \sim E_{|V|}$.
    \item For each optimization step $t = 1, ..., T$:
        \begin{enumerate}
            \item Retrieve the current mini-batch $(X, Y) \subseteq D$.
            \item Project the current embeddings $P$ onto the nearest neighbor $P' = \text{Proj}_E(P)$. 
            \item Calculate the gradient w.r.t. the projected embedding: $g = \nabla_{P'} L_\text{task}(B(P', X_i), Y_i, \theta)$. 
            \item Update the continuous embedding: $P = P - \gamma g$, where $\gamma$ is the learning rate.
        \end{enumerate}
    \item Perform a final projection: $P = \text{Proj}_E[P]$.
    \item Return the learned hard prompt $P$.
\end{enumerate}

The PEZ algorithm combines the advantages of baseline discrete optimization methods and soft prompt optimization by maintaining continuous iterates while projecting them onto the discrete token space during each forward pass. This approach allows for efficient gradient-based optimization while ensuring that the final learned prompt consists of discrete tokens from the model's vocabulary.

\subsection{Greedy Coordinate Gradient (GCG)}
\label{GCG}
The Greedy Coordinate Gradient (GCG) \cite{zou_universal_2023-1} method is an extension of the AutoPrompt algorithm \cite{shin_autoprompt_2020} for optimizing prompts in language models. The key idea behind GCG is to efficiently find promising candidate replacements for each token in the prompt by leveraging gradients w.r.t one-hot token indicators. The method computes the linearized approximation of replacing the $i$-th token in the prompt, $x_i$, by evaluating the gradient:

\begin{equation}
\nabla_{e_{x_i}} L(x_{1:n}) \in \mathbb{R}^{|V|}
\end{equation}

where $e_{x_i}$ denotes the one-hot vector representing the current value of the $i$-th token, $L$ is the loss function, and $|V|$ is the vocabulary size. The top-$k$ values with the largest negative gradient are selected as candidate replacements for token $x_i$. This process is repeated for all tokens $i \in I$, where $I$ is the set of token indices to be optimized. A subset of $B \leq k|I|$ tokens is randomly selected from the candidate set, and the loss is evaluated exactly on this subset. The replacement with the smallest loss is then made. GCG differs from AutoPrompt in that it considers all coordinates for adjustment in each iteration, rather than choosing a single coordinate in advance. This seemingly minor change leads to substantial performance improvements while maintaining the same computational complexity.

\subsection{Autoregressive Randomized Coordinate Ascent (ARCA)}
\label{ARCA}
The method presented in the paper \cite{jones_automatically_2023} formulates an auditing optimization problem to find prompt-output pairs that satisfy a given criterion for large language models. The auditing objective is defined as $\phi: P \times O \rightarrow \mathbb{R}$, where $P = V^m$ is the set of prompts and $O = V^n$ is the set of outputs, with $V$ being the vocabulary of tokens. The optimization problem is formulated as:

\begin{equation*}
\max_{(x,o) \in P \times O} \phi(x, o) \quad \text{s.t.} \quad f(x) = o,
\end{equation*}

where $f: V^m \rightarrow V^n$ is the completion function that maps a prompt $x$ to an output $o$ using the language model's probability distribution $p_{\text{LLM}}$.

To make the optimization problem differentiable, the constraint $f(x) = o$ is replaced with a term in the objective function:

\begin{equation*}
\max_{(x,o) \in P \times O} \phi(x, o) + \lambda_{p_{\text{LLM}}} \log p_{\text{LLM}}(o | x),
\end{equation*}

where $\lambda_{p_{\text{LLM}}}$ is a hyperparameter and $\log p_{\text{LLM}}(o | x) = \sum_{i=1}^n \log p_{\text{LLM}}(o_i | x, o_1, \ldots, o_{i-1})$.

The paper introduces the Autoregressive Randomized Coordinate Ascent (ARCA) algorithm to solve the differentiable optimization problem. ARCA decomposes the objective function into a linearly approximatable term $s_{i,\text{Lin}}$ and an autoregressive term $s_{i,\text{Aut}}$: 

$$s_i(v; x, o) = s_{i,\text{Lin}}(v; x, o) + s_{i,\text{Aut}}(v; x, o) $$
\begin{align*}
s_{i,\text{Lin}}(v; x, o) &:= \phi(x, (o_{1:i-1}, v, o_{i+1:n})) \\
&\quad + \lambda_{p_{\text{LLM}}} \log p_{\text{LLM}}(o_{i+1:n} | x, o_{1:i-1}, v)
\end{align*}
$$s_{i,\text{Aut}}(v; x, o) := \lambda_{p_{\text{LLM}}} \log p_{\text{LLM}}(o_{1:i-1}, v | x).$$

The linearly approximatable term is approximated using first-order approximations at random tokens, while the autoregressive term is computed exactly. ARCA efficiently computes the approximate objective for all tokens in the vocabulary and then exactly computes the objective for the top-k candidates to update the prompt and output tokens iteratively.

\section{Results and Analysis}

The results of the comparative analysis of various trojan detection methods are presented in Table~\ref{tab:results}. The table showcases the performance of each method in terms of Recall and Reverse-Engineering Attack Success Rate (REASR). 

\begin{table}[ht!]
    \centering
    \begin{tabular}{lcc}
        \toprule
        \textbf{Method} & \textbf{Recall} & \textbf{REASR} \\
        \midrule
        PEZ (baseline)  & 0.105           & 0.052          \\
        GBDA (baseline) & 0.116           & 0.056          \\
        UAT (baseline)  & 0.131           & 0.03           \\
        GCG \tablefootnote{\url{https://github.com/llm-attacks/llm-attacks} \cite{noauthor_llm-attacksllm-attacks_2024}}            & 0.109     & 0.068    \\ 
        ARCA \tablefootnote{\url{https://github.com/ejones313/auditing-llms} \cite{jones_ejones313auditing-llms_2024}} & 0.077     & 0.358    \\ 
        GCG (winning team)           & 0.167     & 0.987    \\ 
        \bottomrule
    \end{tabular}
    \caption{Performance comparison of different trojan detection methods. The table presents recall and REASR metrics for each method.}
    \label{tab:results}
\end{table}

During the competition, it was observed that achieving a high REASR score was relatively easy, even using simple black-box evolutionary algorithms to find triggers that force the desired targets. Most participants were able to achieve REASR scores close to 100\%. However, achieving a meaningful Recall score proved to be significantly more challenging. The top scores suggest that the highest Recall scores were around 0.16, assuming near-perfect REASR scores. This level of Recall is no better than a simple baseline of randomly sampling sentences from a distribution similar to the given training prefixes, which would yield Recall scores between 14-17\% due to accidental n-gram matches when computing BLEU similarity.

The difficulty in achieving high Recall scores raises questions about the feasibility of detecting and recovering trojan prefixes inserted into the model, given only the suffixes. It is speculated that there might be mechanisms to insert trojans into models in a way that makes them provably undiscoverable under cryptographic assumptions. While current published work has only demonstrated this for toy models, generalizing the approach to transformers might be achievable. This suggests that the detectability and back-derivability of trojans in the competition may be due to the organizers intentionally making the problem easier than it could be \cite{zygi_adventures_2024}.

Despite the inability to fully solve the problem, working on the competition led to interesting observations about the viability of trojan detection in general and improved techniques for optimizing LLM input prompts concerning differentiable objective functions.

\subsection{Initialization}

One of the teams \cite{straznickas_takeaways_2024} found that the geometry of the finetuned models had an interesting property: let $(p_1, s_1), (p_2, s_2)$ be two trigger-target pairs that were inserted into the model, where $s_1 \neq s_2$. Then, when performing the search for a trigger that forces $s_2$, initializing the search with $p_1$ would make the convergence much faster, even when $p_1, p_2, s_1, s_2$ had no qualitative relation to each other.
        
This property was only discovered during the test phase of the competition. It was exploited in a simple way: $N$ initialization pools are maintained, and the search procedure for some given target is initialized with the contents of one of those pools. The pools are pre-filled with training trojan pairs and get expanded whenever a forcing trigger is successfully found.

\subsection{Filtering}
The output was post-processed to make it more likely to score higher given the specifics of the scoring function.

The search code was run in FP16 precision, which meant that a small fraction of found triggers wouldn't force the target suffix when evaluated in batch mode. To avoid this, a filtering pass is run where targets are generated from the found triggers in batch mode, and all triggers that fail are thrown out.

In the second filtering stage, it is chosen which 20 triggers should be submitted with each target. Triggers $p_i$ are naively dropped if the target already had a trigger $p_j$ with Levenshtein distance $d(p_i, p_j) < T$ for some $T$.

\subsection{Objective Functions for Trojan Prefix Optimization:}

We explored various objective functions for optimizing trojan prefixes, aiming to isolate the intended prefixes from other strings. 

However, none of these objective functions successfully isolated the intended prefixes. Experiments on the development phase competition models showed that the given intended prefixes were not local optima for these objectives, and the optimization algorithm could easily find better prefixes.
    
The organizers attempted to address this issue in the test phase models, and indeed, the test-phase models performed slightly better in this regard. However, the intended prefixes were still not consistently local optima, although finding improved prefixes required more optimization iterations.

\section{Discussion}

In this paper, we have explored the problem of trojan or backdoor attacks on large language models (LLMs). We have focused on the challenges of identifying intended and unintended triggers in the context of the Trojan Detection Competition 2023 (TDC2023). The main points of our discussion are as follows:

\begin{itemize}
    \item \textbf{Intended and Unintended Triggers:} The problem of distinguishing between intended and unintended triggers is a critical aspect of trojan detection in LLMs. Intended triggers are the specific phrases or patterns used by an adversary during the trojan insertion process to activate the malicious behavior. Unintended triggers, on the other hand, are phrases or patterns that accidentally trigger the malicious behavior without being explicitly designed by the adversary. Identifying the intended triggers is crucial for understanding and mitigating the trojan attack.

    \item \textbf{Difficulty of Reverse Engineering Trojans:} Reverse engineering of the intended trojans in practice appears to be a challenging task. In real-world scenarios, where a competent actor has performed trojan insertion and cover-up, the defender may lack crucial information such as the exact list of malicious outputs, known triggers used in training, or white-box access to the base model before fine-tuning. Without these advantages, trojan detection and reverse-engineering could be extremely difficult or even impossible under certain cryptographic hardness assumptions \cite{goldwasser_planting_2022}.

    \item \textbf{Measuring Trojan Insertion Tightness:} One of the teams proposed two ways to quantify the tightness of a trojan insertion \cite{straznickas_takeaways_2024}:
    \begin{enumerate}
        \item By initializing a search procedure with the intended trigger and measuring how much the objective can be improved with local movement. A tightly inserted trojan trigger should be a local optimum, and nearby points should not significantly outperform the intended solution.
        \item By performing a search or optimization for the payload starting from randomly-initialized points and measuring the success rate or time required for success.
    \end{enumerate}
    In the TDC2023 test phase models, we observed that the intended triggers were more likely to be local optima compared to the dev phase models, suggesting a tighter trojan insertion.
    
    \item \textbf{Additional Thoughts:}
    The phenomenon of unintended triggers and the difficulty in distinguishing them from intended triggers raises important questions about the robustness and interpretability of LLMs. It suggests that these models may have inherent vulnerabilities that can be exploited by adversaries, even without explicit trojan insertion. Developing techniques to identify and mitigate such vulnerabilities will be crucial for ensuring the safety and reliability of LLMs in real-world applications.
    
    Furthermore, the potential existence of a well-behaved connecting manifold between trojans is an intriguing finding that warrants further investigation. Understanding the structure and properties of this manifold could provide valuable insights into the inner workings of LLMs and potentially lead to new approaches for trojan detection and mitigation.
    
    Another promising research direction is devising faster trojan detection methods. \cite{zhao_accelerating_2024} uses a smaller draft model to filter unpromising candidates in GCG resulting in a 5.6 times speedup compared to GCG. Having a faster algorithm to investigate adversarial alignment scenarios allows for more thorough research into enhancing the safety of LLMs and improves the practicality of trojan attacks in real-world scenarios.
    
\end{itemize}

\section{Conclusion}
In this paper, we have investigated the problem of trojan or backdoor attacks on large language models (LLMs) in the context of the Trojan Detection Competition 2023 (TDC2023). Our analysis has highlighted the challenges associated with identifying intended and unintended triggers, as well as the difficulty of reverse engineering trojans in real-world scenarios.

The comparative analysis of various trojan detection methods has revealed that achieving high Recall scores is significantly more challenging than obtaining high Reverse-Engineering Attack Success Rate (REASR) scores. The top-performing methods in the competition achieved Recall scores around 0.16, which is comparable to a simple baseline of randomly sampling sentences from a distribution similar to the given training prefixes. This finding raises questions about the feasibility of detecting and recovering trojan prefixes inserted into the model, given only the suffixes.

We have also explored the potential existence of mechanisms to insert trojans into models in a way that makes them provably undiscoverable under cryptographic assumptions. While current published work has only demonstrated this for toy models, generalizing the approach to transformers might be achievable. This suggests that the detectability and back-derivability of trojans in the competition may be due to the organizers intentionally making the problem easier than it could be.

Despite the inability to fully solve the problem, working on the competition has led to interesting observations about the viability of trojan detection in general and improved techniques for optimizing LLM input prompts concerning differentiable objective functions. The phenomenon of unintended triggers and the difficulty in distinguishing them from intended triggers highlights the need for further research into the robustness and interpretability of LLMs.

In conclusion, the TDC2023 has provided valuable insights into the challenges and opportunities associated with trojan detection in LLMs. While the competition has not yielded a complete solution to the problem, it has laid the groundwork for future research in this area. Developing techniques to identify and mitigate vulnerabilities in LLMs will be crucial for ensuring their safety and reliability in real-world applications.

\bibliography{references}

\end{document}